\documentclass[runningheads]{llncs}

\usepackage{eccv}

\usepackage{eccvabbrv}

\usepackage{graphicx}
\usepackage{booktabs}

\usepackage[accsupp]{axessibility}  

\makeatletter
\def\thanks#1{\protected@xdef\@thanks{\@thanks
        \protect\footnotetext{#1}}}
\makeatother

\usepackage{hyperref}

\usepackage{orcidlink}

\usepackage{adjustbox}
\usepackage{multirow}
\usepackage{colortbl}
\usepackage{graphicx}
\usepackage{pifont}
\usepackage{cases}
\definecolor{Gray}{gray}{0.9}
\usepackage[ruled, linesnumbered, vlined]{algorithm2e}
\usepackage{wrapfig}
\usepackage{marvosym}

\begin{document}

\title{Prompt-Based Test-Time Real Image Dehazing: A Novel Pipeline}

\titlerunning{PTTD}

\author{Zixuan Chen\inst{1}\orcidlink{0009-0006-8283-3664} \and
Zewei He\textsuperscript{\Letter}\inst{1,2}\orcidlink{0000-0003-4280-9708}\thanks{\textsuperscript{\Letter} Corresponding author (zeweihe@zju.edu.cn).} \and
Ziqian Lu\inst{1}\orcidlink{0009-0007-3579-9130} \and\\
Xuecheng Sun\inst{1}\orcidlink{0000-0003-3829-1942} \and
Zhe-Ming Lu\inst{1,2}\orcidlink{0000-0003-1785-7847}
}

\authorrunning{Z.~Chen et al.}

\institute{School of Aeronautics and Astronautics, Zhejiang University \and
Huanjiang Laboratory \\
\email{\{zxchen, zeweihe, ziqianlu, xuechengsun, zheminglu\}@zju.edu.cn}
}


\maketitle

\begin{abstract}
Existing methods attempt to improve models' generalization ability on real-world hazy images by exploring well-designed training schemes (\eg, CycleGAN, prior loss). However, most of them need very complicated training procedures to achieve satisfactory results. For the first time, we present a novel pipeline called Prompt-based Test-Time Dehazing (PTTD) to help generate visually pleasing results of real-captured hazy images during the inference phase. We experimentally observe that given a dehazing model trained on synthetic data, fine-tuning the statistics (\ie, mean and standard deviation) of encoding features is able to narrow the domain gap, boosting the performance of real image dehazing. Accordingly, we first apply a prompt generation module (PGM) to generate a visual prompt, which is the reference of appropriate statistical perturbations for mean and standard deviation. Then, we employ a feature adaptation module (FAM) into the existing dehazing models for adjusting the original statistics with the guidance of the generated prompt. PTTD is model-agnostic and can be equipped with various state-of-the-art dehazing models trained on synthetic hazy-clean pairs to tackle the real image dehazing task. Extensive experimental results demonstrate that our PTTD is effective, achieving superior performance against state-of-the-art dehazing methods in real-world scenarios. 
The code is available at \url{https://github.com/cecret3350/PTTD-Dehazing}.
\end{abstract}

\section{Introduction}
\label{sec:intro}
Hazy images often suffer from low contrast, poor visibility, and color distortion \cite{tan2008CVPR}, imposing a negative impact on the downstream high-level vision tasks, such as object detection, image classification, and semantic segmentation.
According to the atmospheric scattering model (ASM) \cite{Narasimhan2002IJCV,Narasimhan2003TPAMI-ASM}, the hazing process is commonly formulated as:
\begin{equation}
	I(x)=J(x)t(x)+A(1-t(x)),
	\label{eq:eq1}
\end{equation}
where $I(x)$ is the observed hazy image and $J(x)$ denotes the clean image of the same scene.
$A$ and $t(x)$ are the global atmospheric light and the transmission map, respectively.

Image dehazing aims to recover the haze-free image from corresponding hazy input, which is a highly ill-posed problem.
Early approaches tend to solve this challenge by introducing various priors, such as Dark Channel Prior (DCP) \cite{He2009CVPR-DCP,he2010TPAMI}, Non-Local Prior (NLP) \cite{berman2016CVPR}, Color Attenuation Prior (CAP) \cite{zhu2015TIP}, etc.
However, real-world hazy images do not always satisfy the priors and artifacts may be introduced under this condition.

\begin{figure}[t]
	\centering
	\includegraphics[width=0.7\columnwidth]{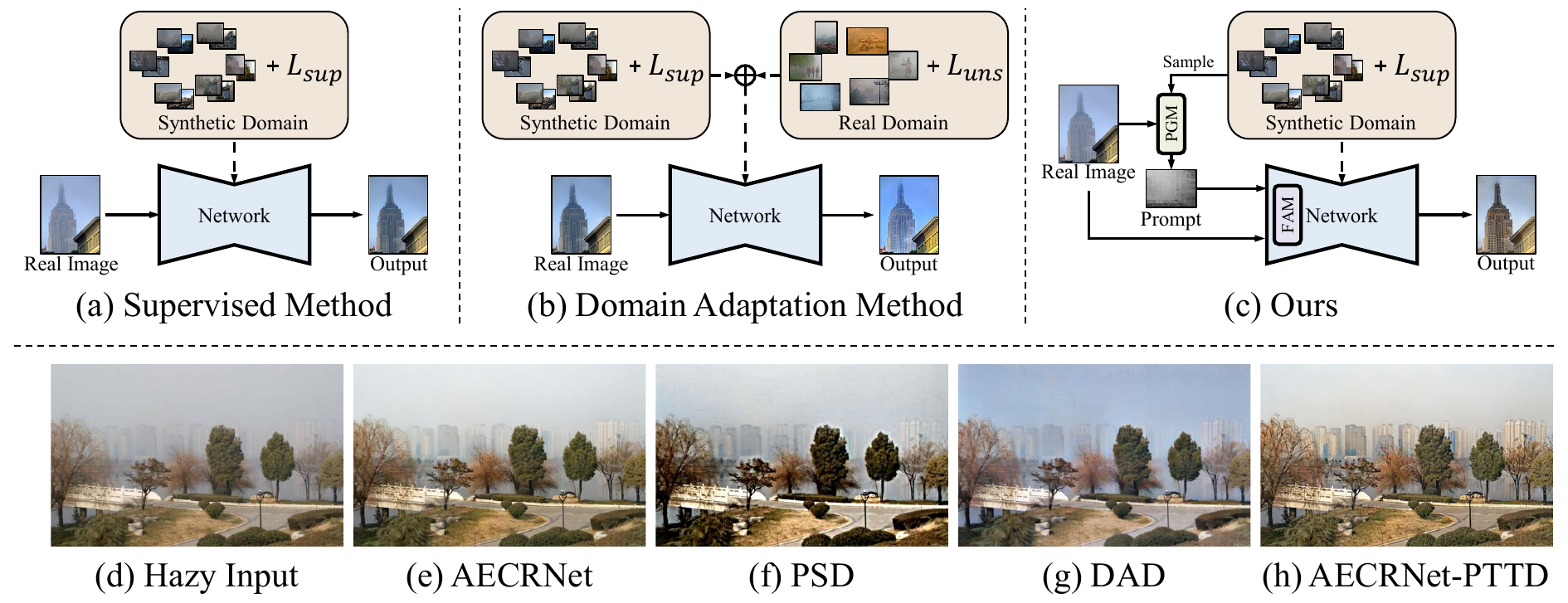} 
	\caption{Different frameworks and their results on real-world hazy images. (a) Directly apply model trained on synthetic data to real hazy images (\eg, AECRNet \cite{wu2021CVPR}); (b) Use synthetic and real data together to train the model and then apply it to real hazy images (\eg, PSD \cite{Chen2021CVPR}, DAD \cite{Shao2020CVPR}); (c) Our proposed PTTD; (d) A real hazy image; and (e-h) Processing results of state-of-the-art (SOTA) methods (AECRNet \cite{wu2021CVPR}, PSD \cite{Chen2021CVPR}, DAD \cite{Shao2020CVPR}) and our proposed PTTD (by adopting pre-trained AECRNet \cite{wu2021CVPR}). It can be observed that AECRNet-PTTD achieves very promising results.} 
	\label{fig:Fig1}
\end{figure}


With the rise of deep learning, researchers have proposed a series of single image dehazing methods based on convolutional neural networks (CNNs).
Most of them try to estimate the $t(x)$ and $A$ \cite{ren2016ECCV,cai2016TIP,li2017ICCV} in \cref{eq:eq1} or directly learn the latent haze-free image (or haze residual) \cite{dong2020CVPR,qin2020AAAI,wu2021CVPR,hong2022AAAI,zheng2023CVPR}.
The former then utilizes the estimated $t(x)$ and $A$ to derive the haze-free image via ASM.

Recently, existing deep learning based dehazing methods are devoted to improving performance by training sophisticated architectures on synthetic datasets \cite{chen2023ARXIV,zheng2023CVPR}.
Though breakthrough progress has been made in the past decade, the processing results on real-captured hazy images are still unsatisfactory (as shown in \cref{fig:Fig1}~(e)), and it is mainly caused by the domain shift.
How to bridge the gap between synthetic and real-world data is a hot research topic among the computer vision community.
In order to improve the models' generalization ability on real-world hazy images, researchers start exploring CycleGAN based \cite{Shao2020CVPR,yang2022CVPR} 
and prior loss based \cite{Golts2020TIP,Chen2021CVPR} methods.
However, either CycleGAN based or prior loss based methods demand very complex training procedures to guarantee the performance.
Moreover, the former tends to produce results with artifacts, whereas the latter is negatively influenced by the inherent deficiencies of physical priors (\cref{fig:Fig1}~(f) and (g)).

\begin{figure}[t]
	\centering
	\includegraphics[width=0.95\columnwidth]{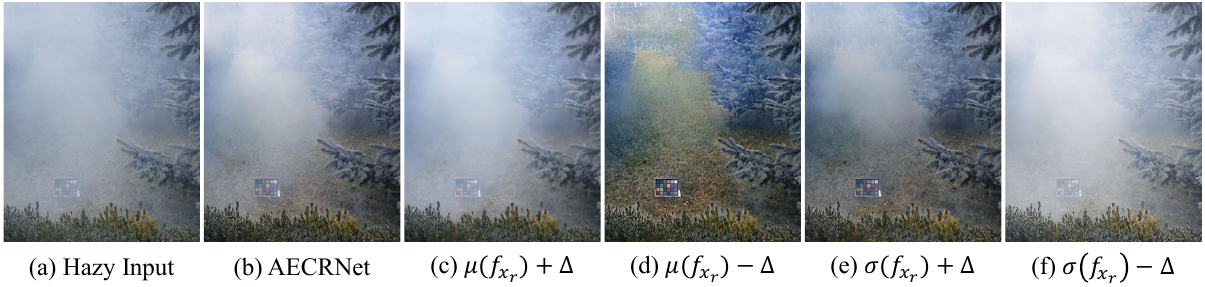} 
	\caption{(a) is a real hazy input; (b) is the processing result of AECRNet \cite{wu2021CVPR}, which is trained on synthetic data; (c) and (d) are the results by fine-tuning the mean $\mu$; (e) and (f) are the results by fine-tuning the standard deviation $\sigma$. In this experiment, the perturbation $\Delta$ is set to a small constant, \ie, $\Delta = 0.005$.}
	\label{fig:Motivation}
\end{figure}

We experimentally find an interesting phenomenon that some statistical indicators (\eg, mean and standard deviation) may significantly affect the model's output, as shown in \cref{fig:Motivation}.
Taking pre-trained AECRNet \cite{wu2021CVPR} as an example, by fine-tuning the mean and standard deviation (only a small perturbation) of deep features extracted by the encoder, the predicted image reconstructed via the decoder is able to manipulate the haze distribution.

Considering this, we propose an effective remedy, called Prompt-based Test-Time Dehazing (PTTD), which attempts to avoid training process and generate visually pleasing results during the inference phase.
Specifically, we first apply a prompt generation module (PGM) to create a visual prompt image, which is later employed to obtain the appropriate values of the statistics.
Similar to style transfer \cite{huang2017ICCV}, the visual prompt accepts a haze-free image as input and inherits the haze distribution from the real hazy image via the operation of image-level normalization (ILN).
Then, we employ a feature adaptation module (FAM) into the encoder of a pre-trained model (PTTD is a plug-and-play pipeline, and can take various dehazing models as its backbone) for adjusting the statistics.
The adjusting values are calculated by taking prompt and real hazy input into consideration, and the adjustment is realized by the operation of feature-level normalization (FLN).
Both of ILN and FLN are derived from adaIN \cite{huang2017ICCV}, which is used to align statistics of the content with those of the style in feature space.
We hope the dehazing results of the real hazy images can better fit human's perception.
To achieve this ultimate goal, the ILN is further revised to avoid color distortion and the FLN adapts features with some effective and mandatory restrictions.
\Cref{fig:pipeline} shows the overall architecture of proposed PTTD.
%


As shown in \cref{fig:Fig1}, we apply the proposed PTTD to AECRNet \cite{wu2021CVPR} by slightly revising original structure (add PGM, FAM). 
We denote it as AECRNet-PTTD.
The AECRNet-PTTD recovers clearer textures, more accurate color, less artifacts than the alternatives, demonstrating the superiority of proposed PTTD.
Our main contributions can be summarized as follows:
\begin{itemize}
	\item A novel dehazing pipeline called Prompt-based Test-Time Dehazing (PTTD) is proposed, which can adapt models pre-trained on synthetic data to real-world images during test-time. To the best of our knowledge, this is the first attempt to tackle the real image dehazing task with \textbf{NO} training process (including training and test-time training).
	\item PTTD aims to narrow the domain gap between synthetic and real by fine-tuning the statistics of encoding features extracted by models pre-trained on synthetic data. PTTD is the first method to explore the correlation between feature statistics and the dehazing effect without modifying the CNN architecture, envisioning an innovative research direction.
	\item We design a Prompt Generation Module (PGM) to obtain a visual prompt via color balanced image-level normalization (CBILN). The prompt provides suitable perturbations for the Feature Adaptation Module (FAM) to adjust the statistics via feature-level normalization (FLN). PGM and FAM constitute our PTTD pipeline, which is simple (no training process), flexible (model-agnostic) and effective (SOTA performance) for real image dehazing.
\end{itemize}



\section{Related Work}
\label{sec:related_works}
\subsection{Single Image Dehazing}
Pioneers of image dehazing~\cite{fattal2008TOG, tan2008CVPR, he2010TPAMI, fattal2014TOG, zhu2015TIP, berman2016CVPR} estimate the key components of ASM (\ie, transmission map and atmospheric light) by leveraging handcraft priors induced from statistics of haze-free images, and then perform image dehazing. 
Although prior-based methods achieve impressive dehazing performance, their effectiveness is constrained to hazy scenes which happen to satisfy the assumption they made.

With the rapid development of deep learning, learning-based methods have dominated the image dehazing domain. 
Early CNN-based methods \cite{cai2016TIP, ren2016ECCV, li2017ICCV, zhang2018CVPR} utilize CNNs to estimate transmission map and atmospheric light. However, the inaccurate estimation of intermediate parameters will introduce the  accumulated error. 
To avoid this, recent CNN-based and transformer-based methods~\cite{liu2019ICCV, dong2020CVPR, dong2020ECCV, qin2020AAAI, wu2021CVPR, guo2022CVPR, hong2022AAAI, ye2022ECCVORAL, song2023TIP, zheng2023CVPR, chen2023ARXIV, He2023Arxiv} discard the physical model and generate dehazing results or their residuals via the end-to-end paradigm.
While learning-based methods demonstrate superior dehazing performance in synthetic domain, they usually suffer from performance degradation on real-world hazy scenes.
\subsection{Real Image Dehazing}
Recently, an increasing number of works have noticed the domain gap between synthetic domain and real domain, and tried to improve real-world dehazing performance. 
One category of these methods utilize real hazy images to reduce the domain gap~\cite{Gandelsman2019CVPR, Golts2020TIP, kar2021CVPR, li2021IJCV, Shao2020CVPR, Chen2021CVPR, yu2022ACMMM, Li_2022_CVPR, Liu2022CVPR-TTT}. DAD ~\cite{Shao2020CVPR} employs CycleGAN for the translation between the synthetic domain and the real domain. PSD~\cite{Chen2021CVPR} fine-tunes pre-trained dehazing models on the real domain by establishing an unsupervised prior loss committee. However, due to the absence of paired data in real-world scenarios, complicate training strategies (\eg, CycleGAN, prior loss) and their corresponding limitations are introduced.
The other category aims to improve generalization performance of dehazing by exclusively utilizing synthetic images~\cite{shyam2021AAAI, yang2022CVPR, wu2023CVPR}.
For example, Wu \etal~\cite{wu2023CVPR} propose a new degradation pipeline to better align the synthetic domain with the real domain. Despite these efforts, without the aid of real hazy images, domain shift remains \cite{wu2023CVPR}.
In order to avoid the inherent limitation of complicate training strategies and leverage the advantages of large-scale synthetic dataset, we propose a novel pipeline to perform feature adaptation on pre-trained models during only test-time. Traditional test-time adaptation (TTA) methods~\cite{yu2022ACMMM,Liu2022CVPR-TTT,yang2024genuine,gou2024tao} in low-level vision adapt models to unlabeled data by updating model parameters. Unlike these works, our method exploits the correlation between intermediate features and the model output, thus eliminating the need for parameter updating.

\section{Methodology}
\label{sec:methods}
Let $D_S=\{x_s^i, y_s^i\}_{i=1}^{N_s}$ denote synthetic hazy-clean pairs and $D_R=\{x_r^i\}_{i=1}^{N_r}$ indicate a collection of real-world hazy images, where $N_s$ and $N_r$ denote the number of synthetic and real-world samples, respectively.
A dehazing network $\mathcal{N}$ trained on $D_S$ often fails to generalize well to real data $D_R$, which can be attributed to the domain gap between $D_S$ and $D_R$.
Our PTTD aims to bridge the gap during the inference phase.

%
%

\begin{figure*}[t]
	\centering
	\includegraphics[width=0.99\textwidth]{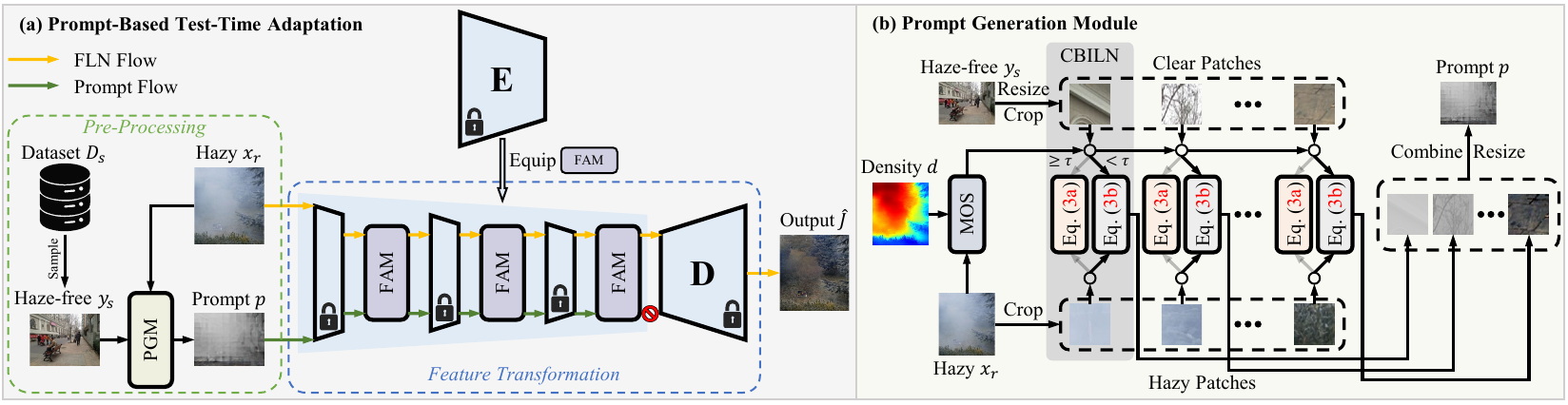} 
	\caption{(a) The overall architecture of our Prompt-based Test-Time Dehazing (PTTD); (b) The proposed prompt generation module (PGM).}
	\label{fig:pipeline}
\end{figure*}

\subsection{Observation and Motivation}
Comparing a hazy image with a haze-free image, the difference is mainly observed in terms of luminance and contrast. 
We argue that these two image-level statistics are correlated closely with feature statistics, \ie, mean $\mu$ and standard deviation $\sigma$.
Adjusting the values of $\mu$ and $\sigma$ in feature space can in turn affect the luminance and contrast of the reconstructed image, achieving haze removal effect.
To validate this hypothesis, we firstly employ a dehazing model $\mathcal{N}$ which is pre-trained on only synthetic hazy-clean pairs (without loss of generality, AECRNet \cite{wu2021CVPR} is adopted here) to perform dehazing on a real hazy input $x_r$.
As shown in \cref{fig:Motivation}~(b), AECRNet achieves unsatisfactory dehazing result.
Then, we apply a subtle perturbation $\Delta$ on $\mu$ and $\sigma$ of the extracted features through AECRNet's encoder.
Let $f_{x_r}$ denotes the features, we adjust $\sigma(f_{x_r})$ to $\sigma(f_{x_r})+\Delta/\sigma(f_{x_r})-\Delta$ via a linear transformation.
As shown in \cref{fig:Motivation}~(e) and (f), $\sigma(f_{x_r})+\Delta$ strengthens $\mathcal{N}$'s dehazing effect, and $\sigma(f_{x_r})-\Delta$ makes the reverse optimization. 
Similar results can be observed in terms of $\mu(f_{x_r})$.

%
%
%
%
%


Motivated by this, we argue a possible solution which can narrow the domain gap is to adjust the statistics of $f_{x_r}$ (with suitable perturbations) to make the features align with the synthetic domain. So the weight-fixed decoder of $\mathcal{N}$ can reconstruct haze-free predictions.


\subsection{Pipeline Overview}
{\color{black}
As illustrated in \cref{fig:pipeline}~(a), PTTD contains two major parts: pre-processing part and feature transformation part.
In the pre-processing part, we employ a prompt generation module (PGM) to produce a visual prompt $p$, whose extracted features $f_p$ are the target domain for guiding the adaption of $f_{x_r}$. 
As the key of our PTTD, feature transformation part can adopt most dehazing models as the backbone.
In the feature transformation part, the feature adaption module (FAM) is employed into the encoder of $\mathcal{N}$ to conduct the adaptation process for narrowing domain gap. 
We denote the new model with FAM as $\mathcal{N}^\dagger$.
$\mathcal{N}^\dagger$ takes $x_r$ and $p$ as inputs and outputs the reconstructed result $\hat{J}$. 
}

%

\subsection{Prompt Generation Module}

Prompt generation module (PGM) is the key component of the pre-processing part.
The goal of PGM is to generate a visual prompt $p$, which can guide the adjustment of statistics inside the FAM (offer appropriate values of mean $\mu$ and standard deviation $\sigma$).
To fulfill this objective, the prompt $p$ must meet two criteria:
(1) it should be similar to the synthetic domain, so the encoder part of a certain pre-trained model can correctly extract the features.
(2) it must have similar haze distribution with $x_r$, so the adaption could be stable to avoid the collapse of the whole pipeline.
These principles incline us to sample a haze-free image $y_s$ from $D_S$, and then synthesize similar haze as $x_r$ on $y_s$. 
For simplicity, the $y_s$ is randomly chosen from the clean images of synthetic hazy-clean pairs.


Adaptive instance normalization (adaIN) \cite{huang2017ICCV} tries to enable arbitrary style transfer by aligning the {\color{black}channel-wise} mean and variance of the content features with those of the style features.
Accordingly, we adopt adaIN by taking $x_r$ and $y_s$ as the style input and content input, respectively.
We denote it as image-level normalization (ILN) by aligning $\mu$ and $\sigma$ of $y_s$ with those of $x_r$:
\begin{equation}
    p=\text{ILN}(y_s, x_r) = \sigma_c(x_r)\frac{y_s - \mu_c(y_s)}{\sigma_c(y_s)} + \mu_c(x_r),
    \label{eq:Eqn.2}
\end{equation}
where $\mu_c(\cdot) \in \mathbb{R}^C$ and $\sigma_c(\cdot) \in \mathbb{R}^C$ are mean and standard deviation, computed across the spatial dimensions for R,G,B channels.

\begin{figure}[t]
	\centering
	\includegraphics[width=0.8\columnwidth]{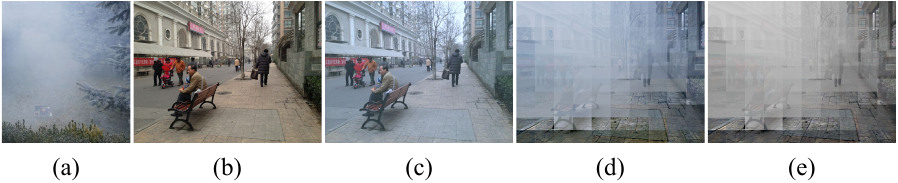} 
	\caption{(a) A real hazy image $x_r$; (b) A haze-free image $y_s$; (c) Prompt via direct ILN (adaIN); (d) Prompt via ILN with partition; (e) Prompt via CBILN with partition.}
	\label{fig:prompts}
\end{figure}

As shown in \cref{fig:prompts}~({\color{black}c), simply} adopting the adaIN scheme would fail to transfer the daze distribution from $x_r$ to $y_s$, since light-haze and dense-haze regions may compromise each other and generate average results.
We argue that $\mu$ and $\sigma$ are global statistics, and the haze distribution (\eg, density) is inconsistent across the whole image.
Therefore, we uniformly crop $x_r$ and $y_s$ into small patches without overlap \footnote{In practice, the resolutions of $x_r$ and $y_s$ may be different, we first resize $y_s$ to the resolution of $x_r$ to improve the flexibility.} and apply ILN on patches.
With this partition strategy, our ILN can effectively transfer the haze distribution from $x_r$ to $y_s$. 
\cref{fig:prompts}~(d) shows the generated prompt, sharing very similar haze with $x_r$.

The criterion (2) is definitely satisfied under this setting.
We surprisingly find that criterion (1) is also met.
We reveal that ILN with partition strategy can be viewed as the process of haze simulation based on ASM.
By comparing \cref{eq:eq1} and \cref{eq:Eqn.2}, if the transmission map $t(x)$ is assumed to be a constant, both equations can be regarded as the linear transformation.
With a sufficient number of partition regions, the transmission map $t(x)$ of a certain patch can be approximately described as a fixed value, which satisfies the key assumption.
In this situation, prompt generated via ILN with partition shares the same characteristics with the synthetic domain, since the prompt strictly follows the ASM.


Inspired by \cite{Li_2022_CVPR}, we also take intra-domain gap (\ie, varicolored hazy scenes) into consideration to calibrate the color distortion.
As illustrated in \cref{fig:prompts}, the real hazy image is often varicolored (\ie, bluish here) and the generated prompt via ILN has similar color distribution.
We argue that the prompt must be color balanced to avoid introducing color distortion in the predicted haze-free image.
There have been many methods developed to provide color constancy, for example, the widely used gray world assumption \cite{buchsbaum1980spatial,Reinhard2001-GWA} hypothesizes that the average scene captured in an image is gray.
We accordingly modify the ILN into color balanced ILN (CBILN) by replacing $\mu_c(x_r)$ and $\sigma_c(x_r)$ with corresponding mean values among R,G,B channels, \ie, $\overline{\mu} = \text{mean}(\mu_c(x_r))$ and $\overline{\sigma} = \text{mean}(\sigma_c(x_r))$.
\begin{subnumcases}{p=\text{CBILN}(y_s, x_r)=}
	\text{ILN}(y_s, x_r), \,\,\,\text{MOS}(x_r) \geq \tau \label{eq:eq3a} &\\
	\overline{\sigma}\frac{y_s - \mu_c(y_s)}{\sigma_c(y_s)} + \overline{\mu}, \,\,\,\text{otherwise} \label{eq:eq3b}&
\end{subnumcases}


Following \cite{dhara2021TCSVT}, we utilize the measure of the spread of hue space (MOS) as our criteria to indicate whether a hazy region (which is determined by haze density $d$ computed via DCP) is varicolored or not.
Smaller MOS value means this image tends to be varicolored and CBILN should be adopted, and vice versa.
As illustrated in \cref{fig:pipeline}~(b), we utilize a threshold $\tau=0.005$ for choosing \cref{eq:eq3a} or Eq.~(\ref{eq:eq3b}) to generate the prompt.

\subsection{Feature Adaptation Module}

As stated before, our PTTD can be easily applied on dehazing models pre-trained on synthetic data.
Typically, we adopt the encoder-decoder-like architecture which is quite prevalent among the pre-trained models \cite{dong2020CVPR,wu2021CVPR,chen2023ARXIV}.
In the feature transformation part, we keep the decoder part unchanged and modify the encoder part by inserting the feature adaptation module (FAM) after the basic blocks in every level
{\color{black}(different spatial sizes indicate different levels).}
Taking the generated prompt $p$ and a real hazy image $x_r$ as inputs, the encoder part of a certain pre-trained dehazing model $\mathcal{N}$ extracts corresponding features of different levels, denoted as $f_{p}^{l}$ and $f_{x_r}^{l}$.
{\color{black}
We omit the level indicator for simplification, and one-level case is presented in this section.
It is not laborious to extend to multiple levels.}
According to the key observation in \cref{fig:Motivation}, a naive idea is to align the {\color{black}channel-wise} statistics of $f_{x_r}$ with those of $f_{p}$ (we denote this as feature-level normalization - FLN), meanwhile make sure the $\sigma_c(f_p)$ is larger than $\sigma_c(x_r)$ and the $\mu_c^+(f_p)$/$\mu_c^-(f_p)$ is smaller than $\mu_c^+(x_r)$/$\mu_c^-(x_r)$.
However, since the visual prompt $p$ is derived from a selected image, the values of $\sigma_c(f_p)$ and $\mu_c(f_p)$ may fluctuate.
We need to add some restrictions on the statistical calculation of $f_p$, and the FLN can be formulated as:
\begin{equation}
	\text{FLN}(f_{x_r}, f_p) = \sigma_c(f_p, f_{x_r})\frac{f_{x_r} - \mu_c(f_{x_r})}{\sigma_c(f_{x_r})}+\mu_c(f_p, f_{x_r}),
\end{equation}
where $\mu_c(f_p, f_{x_r})$ and $\sigma_c(f_p, f_{x_r})$ calculate the statistics of the target domain, providing appropriate perturbations for haze removal.
The calculations are described by the pseudo code (see in \cref{algorithm:FLN}).

%
%
%
\begin{algorithm}[t]
	\caption{Feature-level normalization - FLN}
	\label{algorithm:FLN}
	\LinesNumbered
	\KwIn {$f_p, f_{x_r}$}
	\KwOut {$\hat{f}_{x_r}$ \tcp*{$\hat{f}_{x_r} = \text{FLN}(f_{x_r}, f_p)$}}
	compute $\mu_c(f_p), \mu_c(f_{x_r}), \sigma_c(f_p), \sigma_c(f_{x_r})$ \;
	\tcc{\emph{\textbf{mean adaptation}}}
	\lIf{$\mu_c(f_p) \cdot \mu_c(f_{x_r}) > 0$} {
		$\mu_c(f_p, f_{x_r}) \leftarrow \min(\mu_c(f_p), \mu_c(f_{x_r}))$
	} 
  \lElse{ 
		$\mu_c(f_p, f_{x_r}) \leftarrow \mu_c(f_{x_r})$
	}
	
	\tcc{\emph{\textbf{std adaptation}}}
	\lIf{$\frac{\sigma_c(f_p) - \text{mean}(\sigma_c(f_{x_r}))}{\text{std}(\sigma_c(f_{x_r}))} < \alpha$} {
		$\sigma_c(f_p, f_{x_r}) \leftarrow \max(\sigma_c(f_p), \sigma_c(f_{x_r}))$
	} 
  \lElse{
		$\sigma_c(f_p, f_{x_r}) \leftarrow \sigma_c(f_{x_r})$
	}
	$\hat{f}_{x_r} \leftarrow \sigma_c(f_p, f_{x_r})\frac{f_{x_r} - \mu_c(f_{x_r})}{\sigma_c(f_{x_r})}+\mu_c(f_p, f_{x_r})$
\end{algorithm}

In step 2 and 3, the computed $\mu$ is less than or equal to $\mu_c(f_{x_r})$.
The $\mu_c(f_p) \cdot \mu_c(f_{x_r}) \leq 0$ situation means these two items belong to different quadrants, and in this situation, we abort the feature adaptation from $f_{x_r}$ to $f_p$ to achieve more natural dehazing results.


In step 4 and 5, the computed $\sigma$ is larger than or equal to $\sigma_c(f_{x_r})$.
Although applying a positive perturbation $+\Delta$ on $\sigma$ can yield positive gains for hazy regions, it may be overly aggressive if there is a significant magnitude of difference between the $\sigma_c(f_p)$ and $\text{mean}(\sigma_c(f_{x_r}))$.
The condition $\frac{\sigma_c(f_p) - \text{mean}(\sigma_c(f_{x_r}))}{\text{std}(\sigma_c(f_{x_r}))} < \alpha$ avoids significant fluctuation, restricting the distance from the mean of $\sigma_c(f_{x_r})$.
In our implementation, the hyper-parameter $\alpha$ is set to 2.

\section{Experiments}
\label{sec:experiments}

\subsection{Experimental Configuration}

\noindent\textbf{Datasets.}
Our proposed PTTD pipeline is a test-time approach and model-agnostic, which can be regarded as a basic module to be plugged into most state-of-the-art dehazing models.
Therefore, we don't need training data, and PTTD is free of training.
Moreover, we adopt total six real-world datasets for evaluation.
Four datasets with ground truths (O-HAZE~\cite{ancuti2018CVPRW}, I-HAZE~\cite{I-HAZE}, NH-HAZE~\cite{NH-HAZE} and Dense-Haze~\cite{Dense-Haze}), and two datasets without ground truths (RTTS~\cite{li2018TIP} and Fattal's~\cite{fattal2014TOG}).
O-HAZE, I-HAZE, NH-HAZE and Dense-Haze consist of 45, 35, 55 and 55 pairs of real hazy and corresponding haze-free images captured in various scenes.
RTTS contains over 4000 real hazy images with diverse scenes, and haze-free images are not provided.
In addition, Fattal's dataset with 31 classical real hazy images is also included for evaluation.

\noindent\textbf{Implementation Details.}
We select three state-of-the-art models pre-trained on synthetic data as the backbone to comprehensively show the flexibility of our PTTD pipeline, including a CNN-based (\ie, AECRNet \cite{wu2021CVPR}), a transformer-based (\ie, Dehazeformer \cite{song2023TIP}), and a general image restoration model (\ie, NAFNet \cite{chen2022ECCV}).
For fair comparisons, we re-train these three models on the same data built by Wu \etal~\cite{wu2023CVPR} \footnote{The  models re-trained on Wu's dataset \cite{wu2023CVPR} outperform the original.}, and apply the proposed PTTD to narrow the domain gap from synthetic to real (\textbf{We also perform experiments on their original models in our supplementary materials}).
Typically, we traverse the haze-free images in Wu's dataset \cite{wu2023CVPR} (take it as $y_s$) and calculate the PSNR values (of AECRNet-PTTD model) on exclusive NH-HAZE2 dataset \cite{ancuti2021ntire-NH-HAZE2} to search the top-performing $y_s$ \footnote{NH-HAZE2 dataset is adopted in the famous New Trends in Image Restoration and Enhancement (NTIRE) competition \cite{ancuti2021ntire-NH-HAZE2}, and has no intersection with the testing datasets (including NH-HAZE).}.
The chosen $y_s$ is illustrated in \cref{fig:prompts}~(b).
In PGM, we cropped square patches from resized $y_s$ by setting side length to $\frac{W}{10}$, where $W$ denotes the width of the real hazy input $x_r$.
Another hyper-parameter is the $\alpha$ in \cref{algorithm:FLN}, and we set $\alpha=2$.



\begin{figure*}[t]
	\centering
	\includegraphics[width=0.95\linewidth]{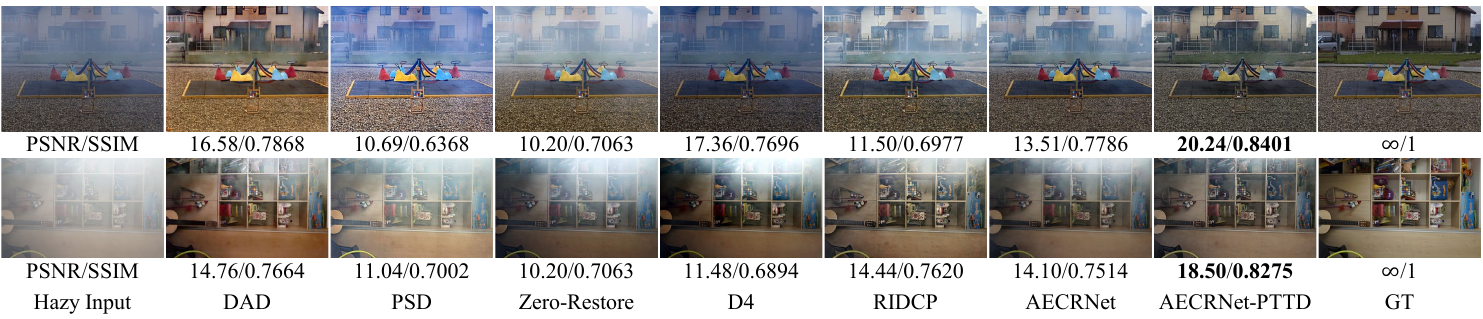} 
	\caption{Dehazing results of various methods on O-HAZE. We choose AECRNet-PTTD to compare with SOTA dehazing methods. Please zoom in on screen for a better view.}
	\label{fig:fig5}
\end{figure*}

\begin{table*}[t]
	\centering
	\caption{Benchmark results of various dehazing methods on O-HAZE, I-HAZE, NH-HAZE, and Dense-Haze datasets. \textbf{Bold} numbers indicate that the proposed PTTD profile can successfully help boost the performance.}
	\adjustbox{width=0.95\linewidth}{
		\begin{tabular}{l|cc|cc|cc|cc}
			\toprule
			\multirow{2}{*}{Method} & \multicolumn{2}{c|}{O-HAZE} & \multicolumn{2}{c|}{I-HAZE} & \multicolumn{2}{c|}{NH-HAZE} & \multicolumn{2}{c}{Dense-Haze} \\
			& PSNR$\uparrow$ & SSIM$\uparrow$ & PSNR$\uparrow$ & SSIM$\uparrow$ & PSNR$\uparrow$ & SSIM$\uparrow$ & PSNR$\uparrow$ & SSIM$\uparrow$\\
			\midrule \midrule
			(TIP'20) DeepDCP & 16.92 & 0.6789 & 14.92 & 0.7171 & 13.04 & 0.4603 & 11.37 & 0.4505 \\
			(IJCV'21) YOLY & 15.83 & 0.6640 & 15.21 & 0.6870 & 12.37 & 0.4561 & 11.60 & 0.4534 \\
			(CVPR'21) Zero-Restore & 16.65 & 0.7536 & 16.56 & 0.7909 & 11.29 & 0.5151 & 12.29 & 0.4411 \\
			\midrule
			(CVPR'20) DAD & 18.36 & 0.7484 & 18.02 & 0.7982 & 14.34 & 0.5564 & 13.51 & 0.4627 \\
			(CVPR'21) PSD & 11.66 & 0.6831 & 13.79 & 0.7379 & 10.62 & 0.5246 & 9.74 & 0.4311 \\
			(CVPR'22) D4 & 16.96 & 0.7229 & 15.64 & 0.7294 & 12.67 & 0.5043 & 11.50 & 0.4500 \\
			(CVPR'23) RIDCP & 16.52 & 0.7154 & 16.88 & 0.7794 & 12.32 & 0.5341 & 9.853 & 0.4525 \\
			\midrule
			(TIP'23) Dehazeformer & 15.23 & 0.7437 & 17.42 & 0.8184 & 11.78 & 0.5423 & 9.25 & 0.4425 \\
			\rowcolor{Gray}(Ours) Dehazeformer-PTTD & \textbf{17.84}$^{+2.61}$ & \textbf{0.7925}$^{+0.0488}$ & \textbf{17.98}$^{+0.56}$ & \textbf{0.8186}$^{+0.0002}$ & \textbf{13.26}$^{+1.48}$ & \textbf{0.5807}$^{+0.0384}$ & \textbf{11.99}$^{+2.74}$ & \textbf{0.4681}$^{+0.0256}$\\
			(ECCV'22) NAFNet & 18.16 & 0.7783 & 17.03 & 0.8095 & 13.17 & 0.5367 & 10.45 & 0.4514 \\
			\rowcolor{Gray}(Ours) NAFNet-PTTD & \textbf{19.63}$^{+1.47}$ & \textbf{0.8094}$^{+0.0311}$ & \textbf{17.94}$^{+0.81}$ & \textbf{0.8280}$^{+0.0185}$ & \textbf{14.08}$^{+0.91}$ & \textbf{0.5765}$^{+0.0398}$ & \textbf{13.23}$^{+2.78}$ & \textbf{0.4787}$^{+0.0273}$\\
			(CVPR'21) AECRNet & 17.24 & 0.7640 & 16.72 & 0.8098 & 12.65 & 0.5326 & 8.59 & 0.4325 \\
			\rowcolor{Gray}(Ours) AECRNet-PTTD & \textbf{20.23}$^{+2.99}$ & \textbf{0.8145}$^{+0.0505}$ & \textbf{18.47}$^{+1.75}$ & \textbf{0.8198}$^{+0.0100}$ & \textbf{14.54}$^{+1.89}$ & \textbf{0.5932}$^{+0.0606}$ & \textbf{13.67}$^{+5.08}$ & \textbf{0.4440}$^{+0.0115}$\\
			\bottomrule
		\end{tabular}
	}
	\label{tab:Tab1}
\end{table*}

\subsection{Experiments on Real-captured Datasets with Labels}
We first employ O-HAZE~\cite{ancuti2018CVPRW}, I-HAZE~\cite{I-HAZE}, NH-HAZE~\cite{NH-HAZE} and Dense-Haze~\cite{Dense-Haze} datasets to evaluate our PTTD profile.
With the labels (\ie, haze-free images), reference-based metrics PSNR and SSIM are utilized to measure the performance.
The quantitative comparisons on these four datasets are summarized in \cref{tab:Tab1}.
The model with PTTD is denoted with suffix `-PTTD'.
We observe that by employing PTTD, robust improvements can be achieved on selected models, proving the effectiveness of this novel pipeline.

In addition, we also compare the models with `-PTTD' suffix with some recent real image dehazing methods (DAD \cite{Shao2020CVPR}, PSD \cite{Chen2021CVPR}, D4 \cite{yang2022CVPR}, and RIDCP \cite{wu2023CVPR}), an unsupervised method (DeepDCP \cite{Golts2020TIP}), and two zero-shot methods (YOLY \cite{li2021IJCV}, Zero-Restore \cite{kar2021CVPR}).
AECRNet-PTTD achieves state-of-the-art performance on O-HAZE, I-HAZE and NH-HAZE.
Specifically, compared with DAD~\cite{Shao2020CVPR}, AECRNet-PTTD achieves 1.87 dB and 0.0661 gains in terms of PSNR and SSIM on O-HAZE.
Note that, the same $y_s$ is utilized for all models and for all datasets. 
We believe better performance can be achieved by searching optimal $y_s$ for each model and each dataset.

We provide some qualitative comparisons of various methods on O-HAZE in \cref{fig:fig5}.
Note that, there is still a significant amount of haze, remaining in the predicted images of \{PSD, Zero-Restore, D4, RIDCP, AECRNet\}.
DAD removes the haze successfully. However, the overall predictions tend to be warm-toned and some high-frequency information is also lost.
Our proposed AECRNet-PTTD restores images with clearer details and less color distortion, which are closer to the ground-truths.
\textbf{More experimental results can be found in the supplementary materials.}


\begin{table*}[t]
	\centering
	\caption{Quantitative comparisons of various dehazing methods on RTTS dataset and Fattal's dataset. \textbf{Bold} numbers indicate the proposed PTTD profile can successfully boost the performance.}
	\adjustbox{width=0.95\linewidth}{
		\begin{tabular}{l|cccc|cccc}
			\toprule
			\multirow{2}{*}{Method} & \multicolumn{4}{c|}{RTTS} & \multicolumn{4}{c}{Fattal's dataset}\\
			& FADE$\downarrow$ & BRISQUE$\downarrow$ & PAQ2PIQ$\uparrow$ & MUSIQ$\uparrow$ & FADE$\downarrow$ & BRISQUE$\downarrow$ & PAQ2PIQ$\uparrow$ & MUSIQ$\uparrow$\\
			\midrule
			\midrule
			Hazy Input & 2.576 & 37.01 & 66.05 & 53.77 & 1.061 & 21.08 & 71.54 & 63.25 \\
			(CVPR'20) DAD & 1.131 & 32.93 & 66.79 & 49.88 & 0.4838 & 29.64 & 71.56 & 58.64 \\
			(CVPR'21) PSD & 1.044 & 22.22 & 70.43 & 52.80 & 0.4161 & 23.61 & 76.02 & 63.04 \\
			(CVPR'22) D4 & 1.406 & 34.52 & 66.84 & 53.57 & 0.4109 & 20.33 & 73.13 & 63.27 \\
			(CVPR'23) RIDCP & 0.9180 & 21.38 & 70.82 & 59.38 & 0.4083 & 20.05 & 74.64 & 66.88 \\
			\midrule
			(TIP'23) Dehazeformer & 1.047 & 21.62 & 69.90 & 58.46 & 0.4399 & 21.70 & 74.72 & 67.15 \\
			\rowcolor{Gray}(Ours) Dehazeformer-PTTD & \textbf{0.7905}$^{-0.2565}$ & \textbf{17.34}$^{-4.28}$ & \textbf{71.45}$^{+1.55}$ & \textbf{61.04}$^{+2.68}$ & \textbf{0.4164}$^{-0.0235}$ & \textbf{19.68}$^{-2.02}$ & \textbf{75.70}$^{+0.98}$ & \textbf{69.52}$^{+2.37}$\\
			(ECCV'22) NAFNet & 1.121 & 26.25 & 70.08 & 58.89 & 0.4183 & 19.80 & 74.12 & 65.91 \\
			\rowcolor{Gray}(Ours) NAFNet-PTTD & \textbf{0.8267}$^{-0.2943}$ & \textbf{22.72}$^{-3.53}$ & \textbf{70.97}$^{+0.89}$ & \textbf{59.79}$^{+0.90}$ & {0.4289$^{+0.0106}$} & \textbf{18.75}$^{-1.05}$ & \textbf{74.65}$^{+0.53}$ & \textbf{66.98}$^{+1.07}$\\
			(CVPR'21) AECRNet & 1.285 & 23.97 & 70.07 & 58.30 & 0.4319 & 21.44 & 74.41 & 66.83\\
			\rowcolor{Gray}(Ours) AECRNet-PTTD & \textbf{0.7120}$^{-0.5740}$ & \textbf{16.63}$^{-7.34}$ & \textbf{72.04}$^{+1.97}$ & \textbf{62.11}$^{+3.81}$ & \textbf{0.3825}$^{-0.0494}$ & \textbf{19.31}$^{-2.13}$ & \textbf{75.01}$^{+0.60}$ & \textbf{67.69}$^{+0.86}$\\
			\bottomrule
		\end{tabular}}
	\label{tab:Tab2}
\end{table*}
	
\begin{figure*}[t]
	\centering
	\includegraphics[width=0.95\linewidth]{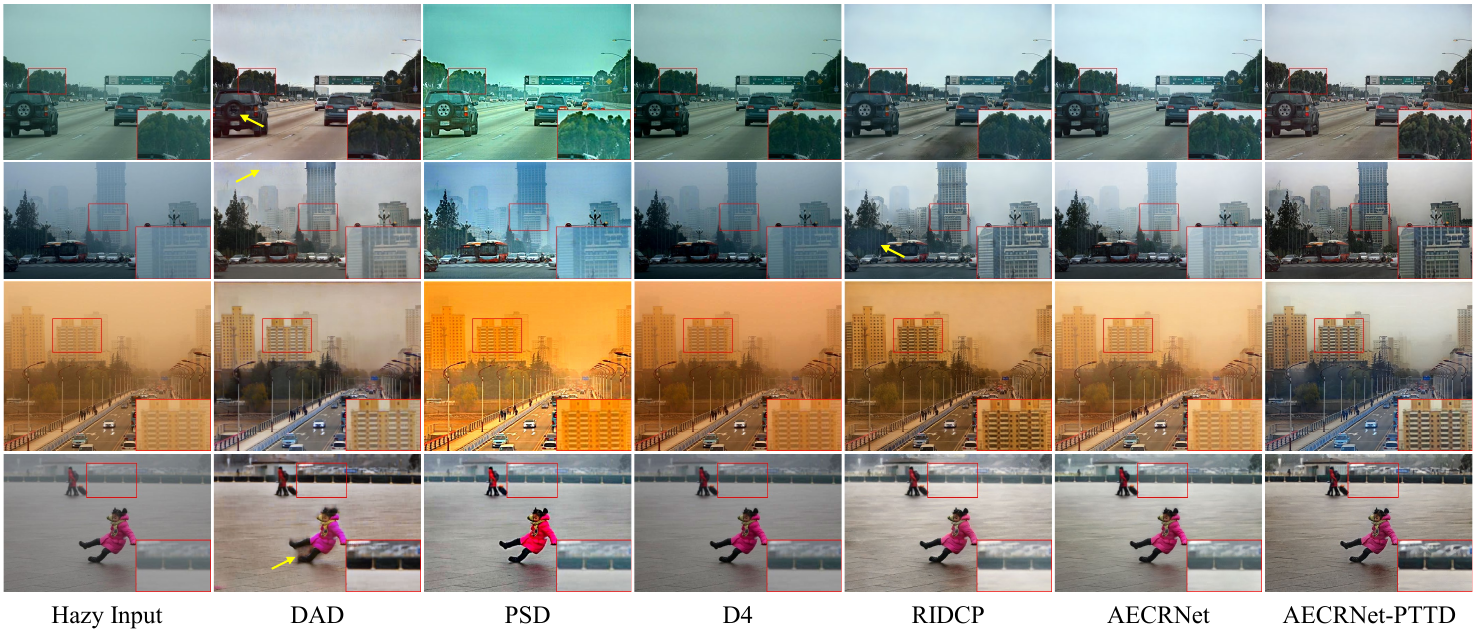} 
	\caption{Dehazing results of various methods on RTTS. We choose AECRNet-PTTD to compare with SOTA dehazing methods. Please zoom in on screen for a better view.}
	\label{fig:fig6}
\end{figure*}

\subsection{Experiments on RTTS and Fattal's}
We then employ RTTS \cite{li2018TIP} and Fattal's \cite{fattal2014TOG} datasets to evaluate our PTTD profile.
Since there is no clean images in these two datasets, we embed some non-reference metrics (\eg, FADE \cite{choi2015TIP}, BRISQUE \cite{mittal2012TIP}, PAQ2PIQ \cite{ying2020CVPR}, and MUSIQ \cite{ke2021ICCV}) to evaluate the quality of dehazing results without reference images~\cite{depictqa_v1}.

The quantitative results are presented in \cref{tab:Tab2}.
We also take the real image dehazing methods \{DAD, PSD, D4, RIDCP\} into the comparison.
With PTTD profile, the selected models achieve robust performance improvements on RTTS.
When it comes to Fattal's dataset, only NAFNet-PTTD fails to boost the performance in terms of FADE.
Note that, RTTS has relatively larger number of images than Fattal's (4332 \textit{vs.} 31).
The metrics measured on latter tend to be unstable and biased.
AECRNet-PTTD achieves state-of-the-art performance on RTTS dataset in terms of all non-reference metrics.
Similarly, the same $y_s$ is utilized for all models and for both datasets.

\Cref{fig:fig6} shows the qualitative comparisons on RTTS.
When it comes to varicolored scenes, all the comparative methods fail to eliminate color cast except DAD.
However, DAD suffers from undesired artifacts, which may cause some distortions or color patches in certain regions.
Only AECRNet-PTTD can restore the visibility and tackle the color shift simultaneously.
In addition, we notice that our PTTD is capable of removing haze for faraway regions in the images, which is usually hard for previous methods.
\textbf{More experimental results can be found in the supplementary materials.}

\begin{wraptable}{r}{0.5\linewidth}
	\caption{Ablation study of PGM and FAM on O-HAZE. The metrics in `Random Prompt' are averaged on all involved $y_s$.}
	\adjustbox{width=\linewidth}{
		\begin{tabular}{c|ccc|cc|cc|cc}
			\toprule
			&\multicolumn{3}{c|}{PGM} & \multicolumn{2}{c|}{FAM} & \multicolumn{2}{c|}{Selected Prompt} & \multicolumn{2}{c}{Random Prompt} \\
			&ILN & CBILN & partition  & adaIN & FLN & PSNR$\uparrow$ & SSIM$\uparrow$ & PSNR$\uparrow$ & SSIM$\uparrow$\\
			\midrule \midrule
			& - & - & - & - & - & 17.24 & 0.7643 & - & - \\
			\midrule
			\ding{172}&- & - & - & \checkmark & - & 16.70 & 0.7638 & 14.58 & 0.7272 \\
			\ding{173}&\checkmark & - & - & \checkmark & - & 16.90 & 0.7777 & 13.92 & 0.7211 \\
			\ding{174}&\checkmark & -& \checkmark & \checkmark & - & 18.28 & 0.7862 & 16.11 & 0.7501 \\
			\ding{175}&- & \checkmark & \checkmark & \checkmark & - & 19.46 & 0.8068 & 17.81 & 0.7826 \\
			\midrule
			\ding{176}&- & \checkmark &\checkmark & - & \checkmark & \textbf{20.23} & \textbf{0.8145} & \textbf{19.51} & \textbf{0.8008} \\
			\bottomrule
	\end{tabular}}
	\label{tab:tab3}
\end{wraptable} 

\subsection{Ablation Study}
In this section, we perform ablation study to verify the effectiveness of (1) PGM, and (2) FAM.
We utilize pre-trained AECRNet \cite{wu2021CVPR} as the backbone and measure the PSNR and SSIM values on O-HAZE dataset.
Since the performance of our PTTD correlates to the haze-free image $y_s$ sampled from $D_S$, we divide ablation experiments into two groups:
(1) selected prompt: elaborately search through the clean images in Wu's dataset \cite{wu2023CVPR} and evaluate the performance on NH-HAZE2 dataset \cite{ancuti2021ntire-NH-HAZE2} to determine the top-performing $y_s$. The selected $y_s$ is then utilized for all experiments.
(2) random prompt: randomly select one clean image from Wu's dataset \cite{wu2023CVPR} and the experimental results are averaged on all involved $y_s$.

\begin{figure}[t]
	\centering
	\begin{minipage}[c]{0.53\textwidth}
		\centering
		\includegraphics[width=\textwidth]{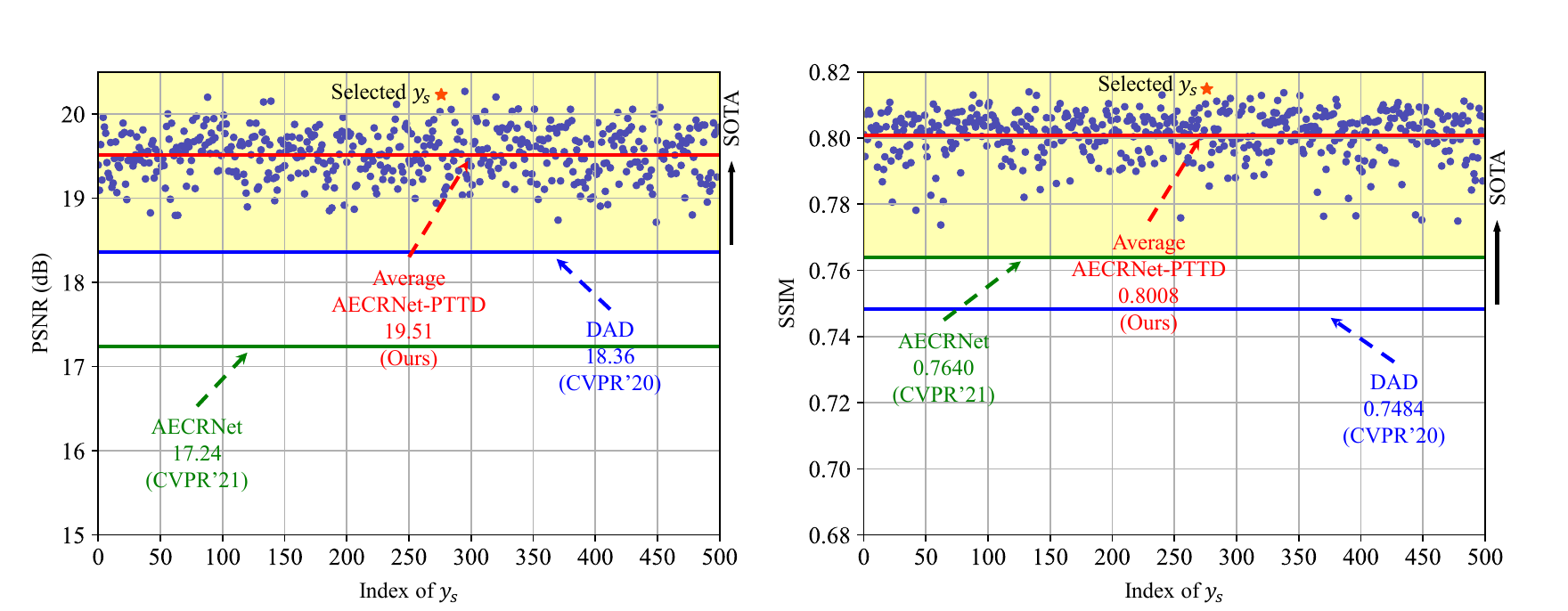}
		\caption{We randomly choose a clean image in Wu's dataset \cite{wu2023CVPR} as $y_s$, and plot the performance scatter graph of each $y_s$ on O-HAZE (using AECRNet-PTTD).}
		\label{fig:fig7}
	\end{minipage} 
	\begin{minipage}[c]{0.46\textwidth}
		\centering
		\includegraphics[width=\textwidth]{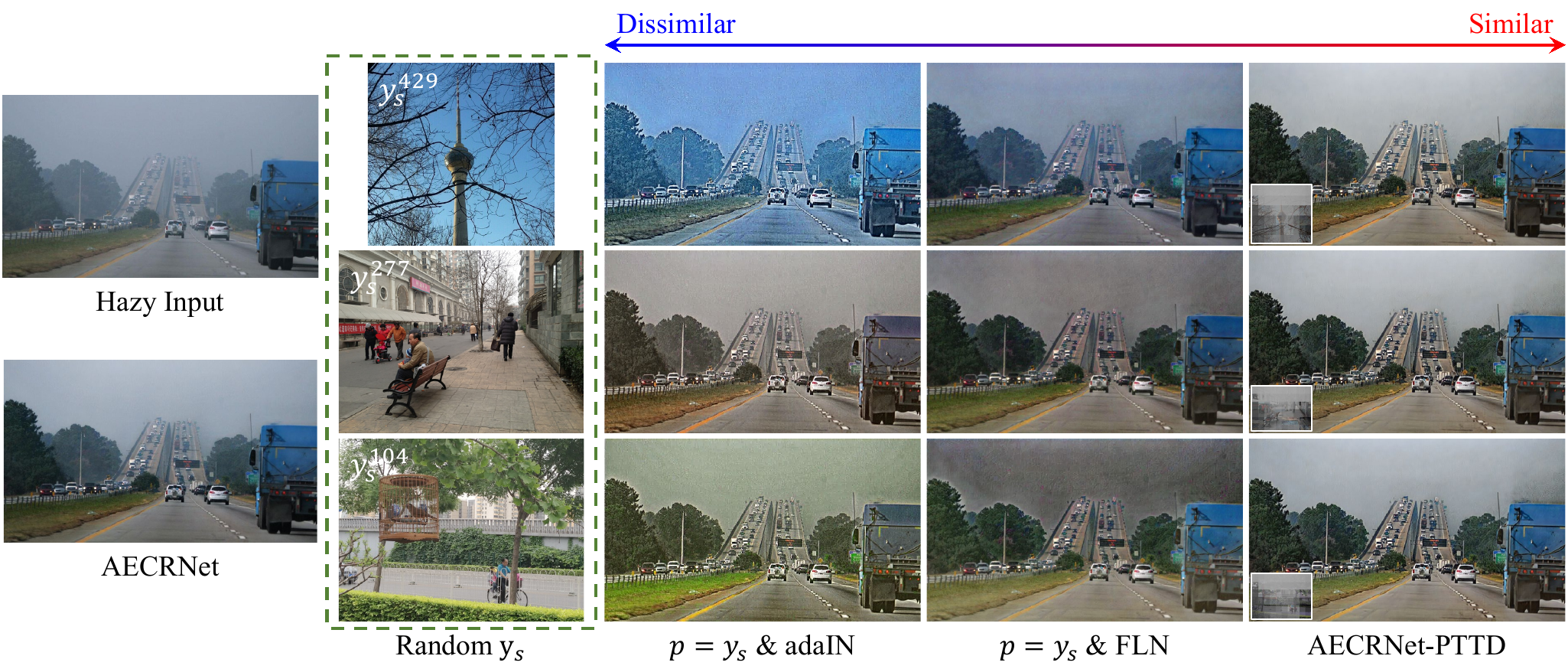} 
		\caption{Visual results of a certain hazy input with varied $y_s$. The superscript of $y_s^{i}$ denotes the index number of Wu's \cite{wu2023CVPR} dataset.}
		\label{fig:ablation}	
	\end{minipage}
\end{figure}

\noindent\textbf{Discussion on the selection of $y_s$.}
The selection of the haze-free image is technically unlimited. 
As shown in \cref{tab:tab3}, though the model using the selected $y_s$ performs better than that with random $y_s$, the latter still reaches state-of-the-art performance.
As illustrated in \cref{fig:fig7}, even the worst-performing $y_s$ from the `Random Prompt' achieves higher PSNR values than DAD and AECRNet.
By grouping 5\% best-performing and 5\% worst-performing clean images, we find that images with rich textures (hierarchical depth information) will fit our PTTD.
When it comes to certain images with large homogeneous region (few textures), the performance somewhat decreases.
Different $y_s$ would lead to different results.
It provides the possibility that the performance can be further improved by searching optimal $y_s$ for every single input.
\textbf{More discussions can be found in the supplementary materials.}
We also experimentally find that by choosing varied $y_s$, the final recovered results are very similar (see in \cref{fig:ablation}), though the intermediate results may be different.
There is no doubt that the model with random $y_s$ works relatively well, indicating the robustness of our proposed PTTD profile.

\noindent\textbf{Effectiveness of PGM.}
The pre-trained AECRNet is regarded as the baseline (the first row of \cref{tab:tab3}).
PGM is designed to generate a visual prompt, and consists of image-level normalization (ILN), color balanced ILN (CBILN), and partition strategy.
To analyze the role of each part, four variants are adopted: 
(1) PGM is not applied and $p = y_s$; 
(2) $p=\text{ILN}(y_s, x_r)$; 
(3) $p=\text{ILN}(y_s, x_r)$ + partition strategy;
(4) $p=\text{CBILN}(y_s, x_r)$ + partition strategy.
The experimental results are summarized in \cref{tab:tab3}.
Directly taking $y_s$ as $p$ or simply adopting ILN to generate $p$ can not help the haze removal and may cause performance drop (16.90 dB and 16.70 dB). 
Partition strategy can guarantee the haze distribution of $p$ to be consistent with $x_r$ and similar to synthetic domain, boosting the performance of ILN to 18.28 dB.
By taking the intra-domain gap into consideration, CBILN pushes the performance forward to 19.46 dB.

\begin{wrapfigure}{r}{0.5\textwidth}
	\centering
	\includegraphics[width=\linewidth]{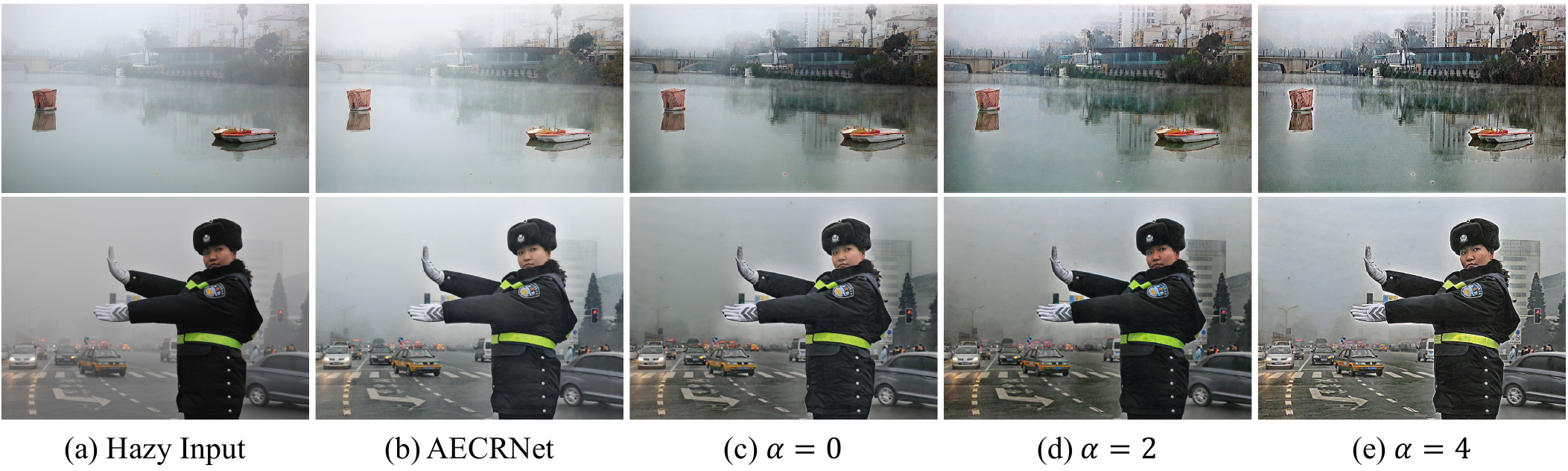}
	\caption{Ablation study of different values of $\alpha$.}
	\label{fig:Fusion}
\end{wrapfigure}

\noindent\textbf{Effectiveness of FAM.}
The function of FAM is to adapt the features $f_{x_r}$ and make them match with the target domain (\ie, $f_{p}$).
We freeze the design of PGM as CBILN with partition, and compare proposed FLN with adaIN.
The main difference is based on the calculations of statistics of the target domain.
The experimental results are summarized in the bottom two rows of \cref{tab:tab3}.
With more restrictions, FLN outperforms adaIN by 0.77 dB.
In addition, we also conduct ablation study on the hyper-parameter $\alpha$, and the qualitative results are illustrated in \cref{fig:Fusion}.
The value of $\alpha$ and the haze removal effect are proportional to each other.
Over-enhanced results may be produced when $\alpha$ is greater than 2, and we can obtain under-enhanced results if $\alpha$ is less than 2.
We set $\alpha$ to 2 to balance the removal effect and visual perception.



\subsection{Inference Time}
\Cref{tab:tab4} demonstrates the inference time comparisons of various dehazing methods.
The results are measured on color images with $512\times 512$ spatial resolution. 
YOLY \cite{li2021IJCV} is time-consuming, since it requires training procedure for every single input.
Due to the calculation procedures of PGM and FAM, PTTD profile inevitably increases the inference time of the adopted backbone.
It is worth mentioning that the increase in inference time of our AECRNet-PTTD against the backbone (\ie, AECRNet \cite{wu2021CVPR}) is acceptable (around 60\%).


\begin{table}[t]
	\centering
	\caption{Inference time comparisons of various dehazing methods.}
	\adjustbox{width=0.6\linewidth}{
		\begin{tabular}{c||c|c|c|c|c|c}
			\toprule
			Method & YOLY \cite{li2021IJCV}  & DAD \cite{Shao2020CVPR} & PSD \cite{Chen2021CVPR} & RIDCP \cite{wu2023CVPR} & AECRNet \cite{wu2021CVPR} & AECRNet-PTTD \\
			\midrule
			\midrule
			\# Param. (M) & 39.99  & 54.59 & 33.11 & 28.72 & 2.611 & 2.611 \\
			Runtime (ms) & 21852  & 19.16 & 29.96 & 173.2 & 23.48 & 38.03 \\
			\bottomrule
		\end{tabular}}
		\label{tab:tab4}
	\end{table}




\section{Limitation and Conclusion}
\label{sec:conclusion}
\begin{figure}[h]
	\centering
	\includegraphics[width=0.95\linewidth]{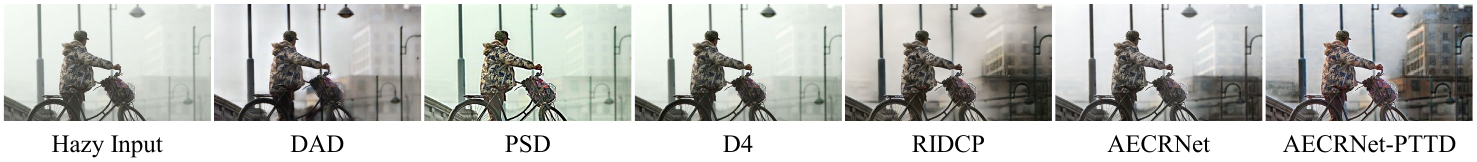} 
	\caption{An imperfect case with artifacts for our PTTD. Our pipeline fails to eliminate the artifacts (\eg, color patches on the building).}
	\label{fig:fig10}
\end{figure}
\noindent\textbf{Limitation.} While our PTTD is capable of generating clean images for distant regions, it may produce artifacts in these areas.
As shown in \cref{fig:fig10}, all the competitors fail to remove the dense haze covering the building, except our AECRNet-PTTD.
However, some artifacts degrade the quality of the recovered result.
Due to limited time, we leave the challenge here and hope future work can address it well.

\noindent\textbf{Conclusion.} In this paper, we propose a PTTD pipeline to adapt models pre-trained on synthetic data to real-world images during the inference phase.
We reveal that fine-tuning the statistics (with suitable perturbations) of encoding features extracted by a certain pre-trained model can help narrow the domain gap between synthetic and real.
Accordingly, a PGM is employed to generate a visual prompt by transferring the haze distribution from the real hazy image to the haze-free image (via the operation of ILN), providing the perturbations for subsequent features adaptation.
Then we modify the encoder of a pre-trained model by adding FAMs to adjust the statistics of extracted features (via the operation of FLN).
Extensive experimental results clearly show the superiority of our PTTD.

\section*{Acknowledgment}
This work was supported in part by National Natural Science Foundation of China under Grant No.~52305590, Zhejiang Provincial Natural Science Foundation of China under Grant No.~LQ24F010004, China Postdoctoral Science Foundation funded project under Grant No.~2022M712792.


%
%

\bibliographystyle{splncs04}
\bibliography{main}
\end{document}